\documentclass[10pt,twocolumn,letterpaper]{article}
\usepackage[pagenumbers]{cvpr} 

\definecolor{cvprblue}{rgb}{0.21,0.49,0.74}
\usepackage[pagebackref,breaklinks,colorlinks,allcolors=cvprblue]{hyperref}
\usepackage{booktabs}
\usepackage{multirow}
\usepackage{siunitx}
\usepackage{graphicx}
\usepackage{caption}
\usepackage{subcaption}
\usepackage{xcolor}
\usepackage{colortbl}
\usepackage{makecell}
\usepackage{natbib}

\def\confName{CVPR}
\def\confYear{2026}

\title{
PhysFire-WM: A Physics-Informed World Model for Emulating
Fire \\
Spread Dynamics
}

\author{Nan Zhou$^1$, Huandong Wang$^2$, Jiahao Li$^1$, Yang Li$^1$, Xiao-Ping Zhang$^1$, Yong Li$^2$, Xinlei Chen$^1$\footnotemark[2]~\\
$^1$Shenzhen International Graduate School, Tsinghua University \\$^2$Department of Electronic Engineering, Tsinghua University\\
}

\begin{document}


\twocolumn[{%
\renewcommand\twocolumn[1][]{#1}%
\maketitle
\begin{center}
    \centering
    \captionsetup{type=figure}
    \includegraphics[width=1\textwidth, trim=0 0 0 0, clip]{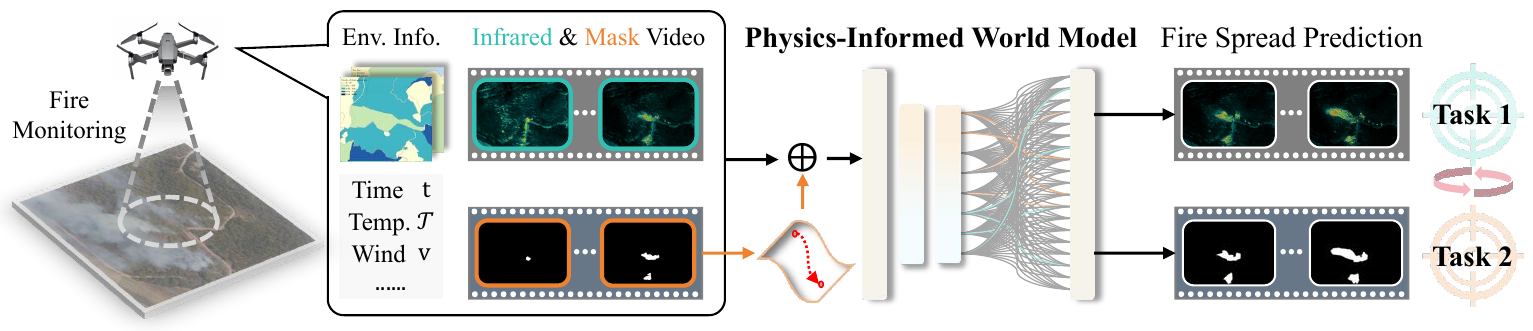}
    \vspace{-15px}
    \captionof{figure}{
    Fire spread modeling via a physics-informed world model. Task 1: Infrared modality prediction. Task 2: Mask modality prediction. “Env. Info.” denotes environmental information.
    }
\label{intro}
\end{center}
}]

\setcounter{footnote}{0}
\renewcommand{\thefootnote}{\fnsymbol{footnote}}
\footnotetext[2]{Corresponding author}


\begin{abstract}

Fine-grained fire prediction plays a crucial role in emergency response. Infrared images and fire masks provide complementary thermal and boundary information, yet current methods are predominantly limited to binary mask modeling with inherent signal sparsity, failing to capture the complex dynamics of fire.
While world models show promise in video generation, their physical inconsistencies pose significant challenges for fire forecasting.
This paper introduces \textbf{PhysFire-WM}, a \textbf{Phy}sics-informed \textbf{W}orld \textbf{M}odel for emulating \textbf{Fire} spread dynamics. 
Our approach internalizes combustion dynamics by encoding structured priors from a Physical Simulator to rectify physical discrepancies, coupled with a Cross-task Collaborative Training strategy (CC-Train) that alleviates the issue of limited information in mask-based modeling. 
Through parameter sharing and gradient coordination, CC-Train effectively integrates thermal radiation dynamics and spatial boundary delineation, enhancing both physical realism and geometric accuracy.
Extensive experiments on a fine-grained multimodal fire dataset demonstrate the superior accuracy of PhysFire-WM in fire spread prediction. Validation underscores the importance of physical priors and cross-task collaboration, providing new insights for applying physics-informed world models to disaster prediction.

\end{abstract}    
\section{Introduction}

\label{sec:intro}


Accurate fire spread prediction is critical for guiding emergency evacuations and directing firefighting efforts~\cite{zhang2025long, ma2025fire}.
As shown in Fig.~\ref{intro}, infrared imagery and fire masks are two key modalities for fire spread characterization, providing thermal-radiation data and spatial-boundary information, respectively~\cite{lutsch2019detection}. However, the nonlinear nature of fire behavior and its environmental interactions pose significant challenges for reliable prediction~\cite{hantson2022human,nowell2025changing,blattmann2025wildfire,reining2025roof,sohel2025world}.

Current fire prediction methodologies are dominated by two prevailing paradigms. Physics-based models simulate fire dynamics through fluid dynamics and heat transfer principles~\cite{meerpoel2023modeling,mcgrattan2010fire,hietaniemi2004fds,mandel2009data,finney1998farsite,mandel2014recent}, yet they exhibit high parameter sensitivity, hindering their adaptation to complex real-world environments. 
Data-driven methods can capture nonlinear features through neural networks~\cite{rajoli2024flamefinder,di2025global,bo2022basnet,lahrichi2025advancing,prapas2022deep,wang2024firevitnet}, yet the majority remain constrained by the sparse signals from the masked modality, leading to limited accuracy due to an inherent information bottleneck.


World models, as a class of generative models designed to understand real-world dynamics, have demonstrated significant potential in video generation, opening new avenues for fire prediction research~\cite{long2025survey}.
However, current models~\cite{wan2025wan,baldassarre2025back,sora2024,bruce2024genie} primarily focus on visual quality while lacking physical constraints, often leading to generated results that violate fundamental physical principles~\cite{zhu2024sora}. 
In fire forecasting, such models may generate physically implausible results like fire fronts propagating upwind or radiation fields violating energy conservation~\cite{motamed2025generative,taketsugu2025physical,bordes2025intphys,foss2025causalvqa}.
When embedded in decision chains, these errors can culminate in severe human and economic losses~\cite{pausas2025role}.

Based on the aforementioned limitations, this study aims to embed multimodal physical knowledge into world models to enhance the reliability of fire prediction, facing two core challenges:
\textit{(C1) Physical Consistency}: how to incorporate physical knowledge described by partial differential equations (PDE) as constraints to ensure the model's outputs adhere to combustion dynamics principles~\cite{feiopen}.
\textit{(C2) Synergistic Modeling}: how to leverage the complementary strengths of the mask modality (spatial boundaries) and the infrared modality (thermal radiation) to achieve enhanced multimodal semantics~\cite{li2024sim2real,gerard2023wildfirespreadts}.


To address these challenges, we propose PhysFire-WM, a physics-informed world model for emulating fire spread dynamics, through two dedicated solutions:
\textit{(S1) To tackle (C1)}, we encode outputs from a Physical Simulator as structured priors, embedding combustion dynamics directly into the generation process through conditional guidance.
\textit{(S2) To tackle (C2)}, we introduce a Cross-task Collaborative Training strategy (CC-Train) that achieves synergistic modeling of thermal dynamics and boundary evolution within a unified framework. Through parameter sharing and gradient coordination, CC-Train leverages cross-modal complementarity to simultaneously enhance physical consistency and enforce geometric precision.

In summary, the main contributions are as follows:
\begin{itemize}
    \item 
    We introduce PhysFire-WM, a physics-informed world model for emulating fire spread dynamics. By unifying physical prior internalization with cross-task collaboration, it delivers both physical plausibility and visual fidelity in complex fire scenarios.
    \item We propose CC-Train, a cross-task collaborative training strategy that bridges infrared and mask prediction tasks through shared parameters and coordinated gradient updates. This strategy exploits inter-modal complementarity to jointly improve thermal distribution consistency and boundary geometric precision.
    \item 
    Experimental results demonstrate that PhysFire-WM achieves state-of-the-art performance in fire spread forecasting. Ablation studies validate the pivotal role of integrating physical priors with cross-task collaboration, providing new insights for physics-informed world models. Code is available in the Supplementary Materials.
\end{itemize}

\section{Related Work}

\subsection{Fire Spread Prediction} 
Existing fire spread modeling approaches can be broadly classified into two paradigms. Physics-based methods simulate fire behavior using principles of combustion and heat transfer~\cite{meerpoel2023modeling,mcgrattan2010fire,hietaniemi2004fds,mandel2009data,finney1998farsite,mandel2014recent}; however, their high parameter sensitivity leads to limited generalizability and predictive accuracy in practical scenarios.
Data-driven approaches, commonly built on UNet~\cite{rajoli2024flamefinder,bo2022basnet,lahrichi2025advancing,garnot2021panoptic}, Transformer~\cite{li2024sim2real,wang2024firevitnet} or LSTM~\cite{huot2020deep} architectures, tend to memorize complex historical patterns while suffering from sparse signal propagation through binary masks. This inherent limitation creates an information bottleneck that fundamentally restricts further improvements in forecasting performance.
In contrast, world models emerge as a promising alternative by mechanistically understanding environmental dynamics~\cite{ding2025understanding}.

\subsection{World Model} 
World models are designed to learn and simulate environmental dynamics for predicting future states~\cite{pausas2025role}. They have achieved remarkable progress in video generation, as evidenced by representative frameworks such as Sora~\cite{sora2024}, Genie~\cite{bruce2024genie}, Cosmos~\cite{agarwal2025cosmos}, Cogvideo~\cite{hong2022cogvideo}, and Wan~\cite{wan2025wan}.
This paradigm aligns naturally with fire evolution prediction, as both tasks require a mechanistic understanding of physical processes such as combustion and the capacity to anticipate dynamic spread trajectories. Current world model architectures~\cite{long2025survey} mainly include recurrent state-space models~\cite{wu2023daydreamer}, diffusion-based models~\cite{blattmann2023stable, ho2022video}, joint-embedding predictive architectures~\cite{assran2023self, assran2025v}, and Transformer-based frameworks~\cite{bruce2024genie, robine2023transformer}. Among these, diffusion transformers (DiT) have attracted growing interest due to their strong performance in preserving temporal coherence and generation quality~\cite{long2025survey}. Building on the DiT framework, this work introduces a physics-informed world model that advances the forecasting accuracy of fire spread dynamics.

\subsection{Physics-aware Generative}
Physics-aware generative modeling has expanded to diverse scenarios~\cite{liu2025generative}. Some approaches enforce explicit physical constraints during training to ensure consistency~\cite{mezghanni2021physically,shang2025roboscape,xie2025physanimator}, yet depend heavily on precise mathematical formulations. Others leverage physics-augmented~\cite{wang2025wisa} or synthetic data~\cite{li2025pisa} to improve performance, though often at considerable computational expense. A third direction embeds physical simulators directly into generative pipelines~\cite{lv2024gpt4motion,yuan2023physdiff}, though such methods often struggle to ensure simulator fidelity and model flexibility.

In parallel, world models have demonstrated significant potential in video generation, yet they have mainly prioritized visual realism while overlooking physical plausibility~\cite{long2025survey,bordes2025intphys}. To bridge this gap, we introduce a unified framework that integrates structured simulation priors with cross-task collaboration, enabling both physically consistent and visually coherent fire spread predictions.
\section{Methodology}


\subsection{Preliminaries}
\label{pre}
\paragraph{PDE Model.}
The dynamics of fire spread are governed by PDEs that capture essential physical mechanisms such as heat diffusion, chemical reactions, and convective transport, thereby establishing a mathematical foundation for physics-based simulators. Central to this system is the thermal balance equation~\cite{mandel2009data, mandel2014recent}, expressed as:
\begin{equation}
c \frac{\partial \mathcal{T}}{\partial t}
= \nabla \cdot (k \nabla \mathcal{T})
- (\vec{v} + \gamma \nabla z) \cdot \nabla \mathcal{T}
+ \underbrace{A F r(\mathcal{T}) - C \Delta\mathcal{T}}_{S(\mathcal{T})},
\label{PDE}
\end{equation}
where $\mathcal{T}$ represents the combustion boundary, $p = (p_1, p_2)$ denotes spatial coordinates, $\nabla = (\partial / \partial p_1, \partial / \partial p_2)$ is the spatial gradient operator, $\vec{v}$ is the wind velocity field, and $F$ is the fuel concentration. The combustion rate $r(\mathcal{T})$ depends on the boundary state, while $\gamma \nabla z$ captures terrain-induced acceleration effects. Physical parameters include heat capacity $c$, thermal conductivity $k$, terrain coefficient $\gamma$, reaction coefficient $A$, and cooling coefficient $C$.


\textit{Physical Interpretation}. 
Each term in Eq.~(\ref{PDE}) carries distinct physical significance: the unsteady term $c \frac{\partial \mathcal{T}}{\partial t}$ describes the temporal evolution of the combustion boundary; the diffusion term $\nabla \cdot (k \nabla \mathcal{T})$ models boundary propagation through radiation and turbulent mixing; the advection term $(\vec{v} + \gamma \nabla z) \cdot \nabla \mathcal{T}$ captures convective transport driven by wind and terrain slope; and the heat source term $S(\mathcal{T})$ represents net combustion effects. 

In Sec.~\ref{compoments}, we develop a physical simulator to numerically solve this PDE system.

\paragraph{Diffusion Transformer.}
We adopt the Wan architecture~\cite{wan2025wan}, a DiT-based framework consisting of three core components.
\textit{Wan-VAE Encoder} compresses input video sequences $V \in \mathbb{R}^{(1+T)\times H\times W\times 3}$ into latent representations $x \in \mathbb{R}^{(1+T/4)\times H/8\times W/8\times 3}$, preserving spatiotemporal integrity while significantly reducing computational complexity.
The \textit{DiT backbone} follows a three-stage design~\cite{peebles2023scalable} with three core modules:
patch embedding partitions inputs into spatiotemporal tokens, transformer blocks model contextual dependencies and integrate conditional signals via multi-head cross-attention, and patch recovery reconstructs outputs to their original resolution while maintaining structural consistency. \textit{Text Encoder} utilizes umT5 with multilingual encoding capabilities~\cite{chung2023unimax}, provides semantically rich and stable text representations, offering precise guidance throughout the diffusion trajectory.

The Wan architecture adopts flow matching techniques to achieve unified modeling of denoising diffusion across both image and video domains~\cite{esser2024scaling}. This formulation enables stable training of continuous-time generative models through ordinary differential equations. 
Given a latent representation $x_1$ and Gaussian noise $x_0\sim\mathcal{N}(0,I)$, we sample a timestep $n\in[0,1]$ from a logit-normal distribution and linearly interpolate between the endpoints: 
\begin{equation}
x_n = n x_1 + (1-n) x_0.
\end{equation}

The ground-truth velocity field is defined as $u_n=x_1-x_0$, and the model is trained to predict this field conditioned on noisy inputs, text embeddings $c_\text{text}$, and the timestep. The resulting loss is:
\begin{equation}
\mathcal{L} = E_{x_0,x_1,c_\text{text},n} \| u_\theta(x_n,c_\text{text},n) - u_n \|^2,
\label{dit-loss}
\end{equation}
where $\theta$ represents model parameters. To improve training efficiency, we fine-tune the DiT backbone using Low-Rank Adaptation (LoRA)~\cite{hu2022lora}, which substantially reduces computational cost while preserving model capacity.

\paragraph{All-in-One Video Generation.}

The development of unified multimodal video generation frameworks is inherently challenged by the need to maintain spatiotemporal dynamic consistency. In response, VACE~\cite{jiang2025vace} proposes an integrated architecture with two core components: a Video Condition Unit (VCU) and a Context Adapter.

\textit{VCU.} 
The VCU integrates three modalities: text prompts $T$, video sequences $V=\{f_1,f_2,...,f_t\}\in R ^{T\times H \times W\times 3}$, and binary mask sequences $M=\{m_1,m_2,...,m_t\}\in \{0_{H \times W},1_{H \times W}\}^{T\times H \times W}$, where mask values 0 and 1 indicate regions to be preserved and modified, respectively. This is formalized as:
\begin{equation}
VCU=[T;V;M]. 
\label{vcu}
\end{equation}

\textit{Context Adapter.} 
This module integrates the VCU into the DiT backbone via a three-stage encoding pipeline: 
(1) Concept Decoupling decomposes input frames into reactive segments $V_c=V\times M$ (for editing) and inactive frames $V_k=V\times(1-M)$ (for keeping), guided by the binary mask $M$; (2) Context Encoding projects $V_c$, $V_k$, and $M$ into a structured latent representation using a pre-trained VAE encoder; (3) Feature Embedding concatenates the encoded features and projects them into context tokens, where $V_c$ and $V_k$ reuse original video embedding weights, while mask embeddings are zero-initialized.

\begin{figure*}[ht]
  \centering
  \includegraphics[width=1\linewidth]{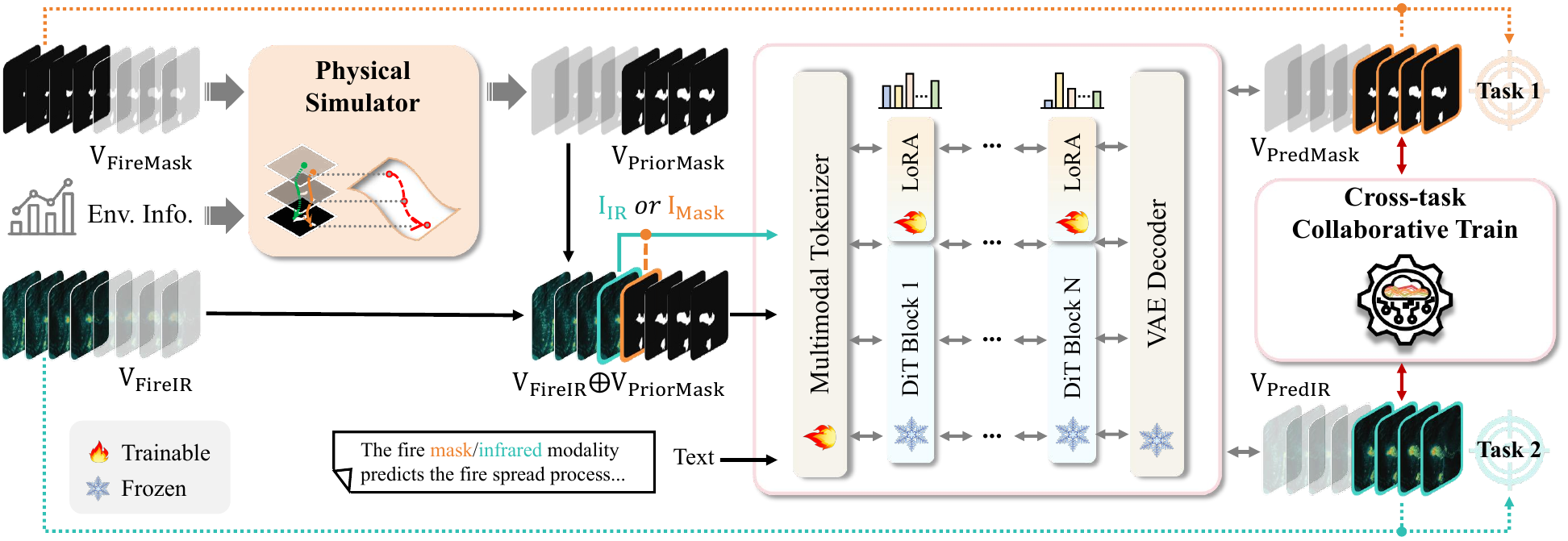}
  \caption{\textbf{Overview of PhysFire-WM.} 
  The pipeline comprises: physical prior generation from the Physical Simulator; unified spatiotemporal tokens production through the Multimodal Tokenizer; joint optimization of infrared and mask prediction via Cross-task Collaborative Training.
  }
  \label{PhysFire-WM}
\end{figure*}

\subsection{Problem Formulation}
\label{single-task}

The objective of fire spread prediction is to learn a mapping function $g$ that infers the dynamic boundary of a fire front, which can be naturally characterized by fire masks. This function takes historical mask observations $V_\text{FireMask}$  and environmental information $\mathcal{E}$  as input, and outputs the predicted mask sequence $V_\text{PredMask}$ at future time steps (see Supplementary Sec.~\ref{sec-note} for the full notation):
\begin{equation}
V_\text{PredMask}=g(V_\text{FireMask}, \mathcal{E}).
\end{equation}

When introducing the world model $\mathcal{W}$  to solve this task, we observe that relying solely on binary mask information limits the model's capacity to understand dynamic processes. To address this, we introduce the infrared modality to provide richer thermal radiation dynamics.
However, since our ultimate goal is to generate mask modality outputs characterizing the dynamic boundary of a fire front, which can be viewed as a domain transferred representation~\cite{hou2021visualizing,chiu2025abc} from the infrared modality.
This naturally leads to two modeling approaches:

\textit{(1) In-Domain Prediction}: The world model $\mathcal{W}$ takes infrared observations $V_\text{FireIR}$, environmental information $\mathcal{E}$, and an infrared prompt $T_\text{IR}$ as input, and outputs the predicted infrared video $V_\text{PredIR}$. Subsequently, an external segmentation model $s$ (e.g., SAM2~\cite{ravi2024sam}) processes the infrared prediction to generate the final mask $V_\text{PredMask}$:
\begin{equation}
V_\text{PredIR}=\mathcal{W}(V_\text{FireIR}, \mathcal{E}, T_\text{IR}),\ V_\text{PredMask}=s(V_\text{PredIR}).
\label{n-d}
\end{equation}

\textit{(2) Cross-Domain Translation}: The world model $\mathcal{W}$ takes infrared observations $V_\text{FireIR}$, environmental information $\mathcal{E}$, and a mask prompt $T_\text{Mask}$ as input, and directly outputs the predicted fire mask $V_\text{PredMask}$. This approach enables end-to-end mask prediction without external modules:
\begin{equation}
V_\text{PredMask}=\mathcal{W}(V_\text{FireIR}, \mathcal{E}, T_\text{Mask}).
\label{d-t}
\end{equation}

\begin{figure*}[ht]
  \centering
  \begin{subfigure}{0.35\linewidth}
    \centering
    \includegraphics[width=\textwidth]{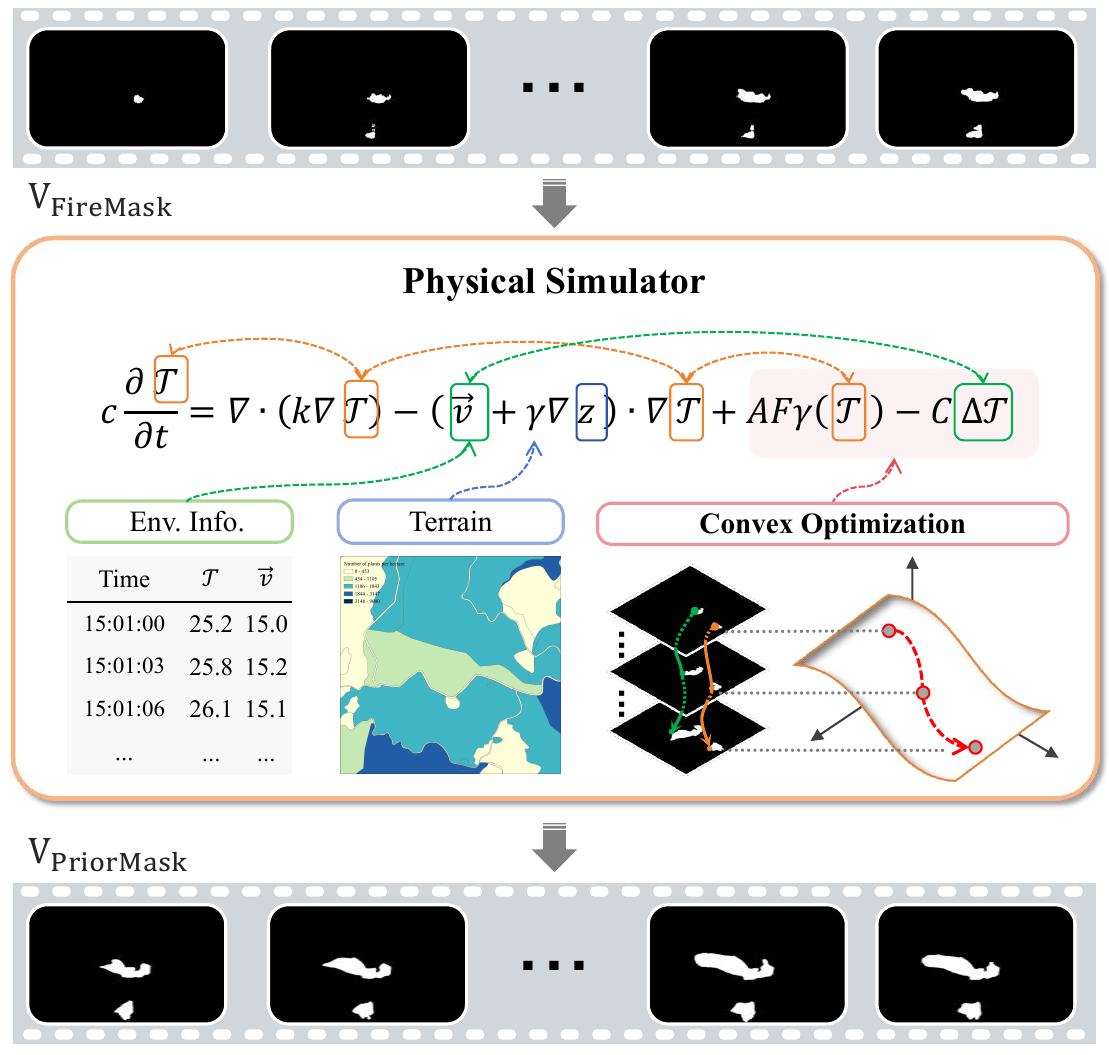}
    \caption{Physical Simulator.}
    \label{sim}
  \end{subfigure}
  \hfill
  \begin{subfigure}{0.63\linewidth}
    \centering
    \includegraphics[width=\textwidth]{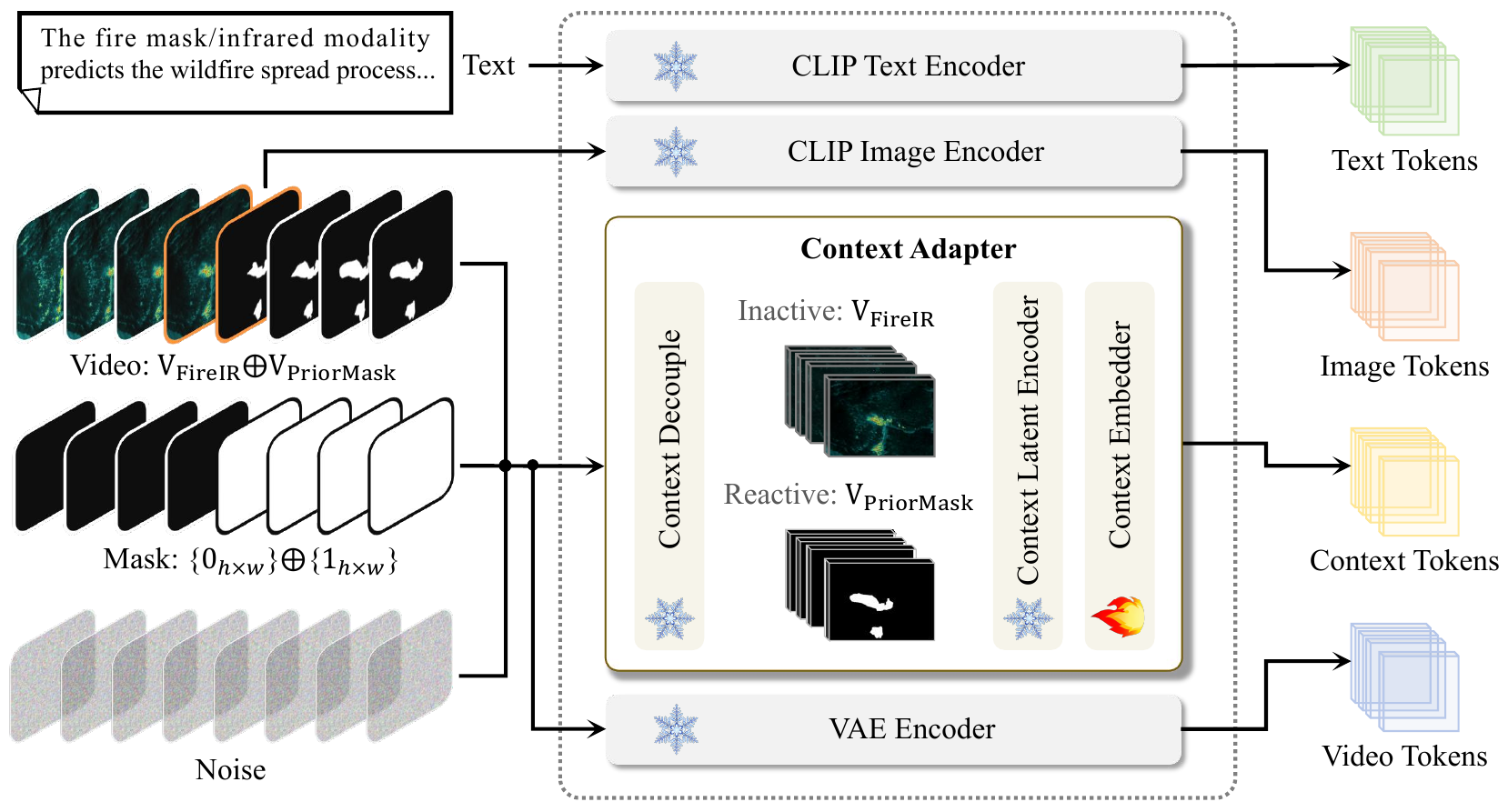}
    \caption{Multimodal Tokenizer.}
    \label{multi}
  \end{subfigure}
  \caption{Components of PhysFire-WM. 
  (a) The Physical Simulator derives physical prior knowledge from PDEs.
  (b) The Multimodal Tokenizer unifies multimodal inputs into spatiotemporally consistent tokens.
  }
  \label{fig:framework}
\end{figure*}

\subsection{PhysFire-WM}
\label{compoments}
We propose PhysFire-WM, a physics-informed world model for emulating fire spread dynamics. As illustrated in Fig.~\ref{PhysFire-WM}, the framework integrates three core components: 
a Physical Simulator $P_{\phi}$, a DiT $G_{\psi}$ enhanced with a Multimodal Tokenizer $E_{\eta}$, and a Cross-task Collaborative Training strategy (CC-Train).

Guided by the fundamental principles of combustion dynamics, we encode the evolutionary patterns revealed by the Physical Simulator $P_{\phi}$ into structured physical priors, which are deeply embedded into the DiT $G_{\psi}$ through the Multimodal Tokenizer $E_{\eta}$. We further propose the CC-Train, establishing a synergistic enhancement mechanism that bridges thermal distribution consistency and boundary geometric precision, thereby achieving a unified framework of physical authenticity and visual fidelity in fire prediction.



\paragraph{Physical Prior Embedding.}
While video generation technology has made remarkable progress, the prevalent lack of explicit physical constraints often leads to generated results that violate fundamental physical laws~\cite{wan2025wan,baldassarre2025back,sora2024,bruce2024genie}.
In fire modeling, for instance, free generation may produce fire fronts propagating upwind or thermal radiation patterns that contradict energy conservation principles.

To bridge this gap, we first design a Physical Simulator $P_{\phi}$ that converts the spatiotemporal evolution laws of combustion dynamics described by partial differential equations into physical priors embeddable into the generation process. Furthermore, we integrate these physical priors with multimodal information to construct a structured conditional module, effectively guiding the world model's generation process to ensure physical consistency and spatiotemporal coherence. The implementation details are as follows:

\textit{Physical Simulator.} 
Specifically, we design a physics-based simulator $P_{\phi}$  (Fig.~\ref{sim}) grounded in the fire energy-transfer partial differential equation (Eq.~(\ref{PDE})). Its operational procedure consists of two main components: (1) since the source term $S(\mathcal{T})$ varies with vegetation type, we model this vegetation-dependent term through convex optimization, representing it as a constrained combination of historical observations; (2) building upon the results of this convex optimization, we compute the fire spread boundary using observed masks and environmental parameters via the finite difference method.
The simulator's final output is expressed as:
\begin{equation}
V_\text{PriorMask}=P_{\phi} (V_\text{FireMask},\mathcal{E}).
\label{phys-prior}
\end{equation}

Complete derivation details of the Physical Simulator are provided in Supplementary Material Sec.~\ref{detail-PS}.

\textit{Physical Prior as Conditional Guidance.}
Based on the physical prior $V_\text{PriorMask}$ obtained from Eq.~(\ref{phys-prior}), we customize the VCU (Eq.~(\ref{vcu})) for the fire prediction task with the following structured input:
\begin{equation}
\begin{aligned}
V &= V_\text{FireIR} \oplus V_\text{PriorMask} = \{r_1, r_2, \dots, r_a, f_1, f_2, \dots, f_b\}, \\
M &= \{0_{H \times W}\} {\times a} \oplus \{1_{H \times W}\} {\times b},
\end{aligned}
\label{fire-vcu}
\end{equation}
where $\oplus$ denotes concatenation along the temporal dimension, $r$ denotes the real infrared frames, and $f$ denotes the mask frames. Accordingly, all-zero masks preserve the infrared content from $V_\text{FireIR}$, while all-one masks designate regions to be reconstructed from $V_\text{PriorMask}$.

The customized VCU (Eq.~(\ref{fire-vcu})) incorporates multimodal fire data as conditional prompts, steering the world model's generation through dual complementary mechanisms:
In the conditional diffusion pathway, the physical prior serves as a structured spatiotemporal constraint, confining fireline evolution to the physically plausible phase space defined by governing partial differential equations. This explicit constraint effectively suppresses error accumulation typically encountered in autoregressive generation.

Simultaneously, in the feature interaction dimension, the physical prior engages with observational modalities through cross-attention mechanisms. This enables the model to maintain generation flexibility while consistently adhering to fundamental physical principles including energy conservation and fire propagation dynamics.

This hybrid explicit-implicit guidance framework addresses key limitations of purely data-driven approaches, such as boundary ambiguity and physical inconsistencies, while achieving dynamic integration of physical principles with observational data through differentiable modeling. The resulting system demonstrates significant improvements in long-term prediction performance, delivering enhanced physical consistency and spatiotemporal coherence in fire spread forecasting.

\paragraph{
Cross-task Collaborative Learning.
}
In fire spread forecasting, infrared and mask modalities provide complementary physical insights: infrared imagery captures detailed thermal radiation distributions, while masks delineate precise spatial boundaries. However, effectively fusing these dual modalities to achieve comprehensive modeling of dynamic fire behavior remains an open challenge. 

To bridge this gap, we first design a Multimodal Tokenizer $E_{\eta}$ that guides the generation process of DiT $G_{\psi}$ toward physically plausible outputs, and subsequently propose two learning tasks derived from the modeling approaches (Sec.~\ref{single-task}) to implement a Cross-task Collaborat Training (CC-Train), which facilitates knowledge sharing and complementary information exchange through coordinated joint optimization. Implementation details are provided below:

\textit{Multimodal Tokenizer.} 
World models were initially designed for video generation from text or single images~\cite{sora2024, bruce2024genie}. Although subsequent studies have attempted to incorporate trajectories or physical vectors as conditional inputs~\cite{xie2025physanimator,lv2024gpt4motion}, existing architectures face key adaptation bottlenecks: missing cross-modal fusion for transient-steady feature alignment; and general encoders that blur modality specifics. These constraints hinder effective adaptation to our infrared-mask dual-stream framework.

To overcome these limitations and achieve comprehensive understanding of customized multimodal inputs (Eq.~(\ref{fire-vcu})), we designed a Multimodal Tokenizer $E_{\eta}$. 
Building upon the standard DiT architecture~\cite{peebles2023scalable}, our design incorporates learnable context adapters (Sec.~\ref{pre}) that establish dedicated pathways for heterogeneous information fusion. This enables simultaneous parsing and differentiation between historical infrared observations and physical prior masks, as illustrated in Fig.~\ref{multi}.

Specifically, the Multimodal Tokenizer $E_{\eta}$ jointly encodes infrared observations $V_\text{FireIR}$, physical priors $V_\text{PriorMask}$, text prompts $T$, image prompts $I$, noisy videos $\mathcal{N}$, and control masks $M$ to generate a semantic token sequence with unified spatiotemporal dimensions:
\begin{equation}
c_\text{Tokens}=E_{\eta} (V_\text{PriorMask},V_\text{FireIR}, M,T,I,\mathcal{N}).
\end{equation}

\textit{DiT.} 
The unified token representations $c_\text{Tokens}$ generated by the Multimodal Tokenizer serve as conditioning inputs to the DiT $G_{\psi}$. 
To adapt to our fire prediction task, we modify the original DiT loss function (Eq.~(\ref{dit-loss})) as follows:
\begin{equation}
\mathcal{L}=E_{x_0,x_1,c_\text{Tokens},n}\|\hat{u}(x_n,c_\text{Tokens},n;\theta)-u_n\|^2,
\end{equation}
where the conditioning tokens $c_\text{Tokens}$ coordinate gradient propagation for dual-modality outputs through the denoising network. 
Together, $P_{\phi}$, $E_{\eta}$, and $G_{\psi}$ integrate to form the complete architecture of PhysFire-WM.

\textit{CC-Train.} 
Existing approaches have consistently failed to effectively integrate the complementary strengths of infrared observations and mask data: the former captures fine-grained thermal radiation distributions, while the latter delineates precise spatial boundaries. This shortcoming is clearly reflected in the two dominant prediction paradigms: native-domain prediction (Eq.~(\ref{n-d})) suffers from cascaded error propagation and depends critically on external models, whereas domain-transformed prediction (Eq.~(\ref{d-t})) struggles to capture complex fire dynamics owing to sparse supervision from binary masks and constrained information flow.

To address these challenges, we introduce CC-Train, a collaborative training mechanism within the PhysFire-WM framework. This approach establishes bidirectional knowledge transfer between infrared and mask modalities, enabling thermal features to guide boundary evolution while spatial constraints direct thermal field reconstruction. Through this process, we achieve deep complementarity and synergistic improvement of both modalities under a unified optimization framework.


The training process utilizes temporally concatenated infrared videos $V_\text{FireIR}$ and physical prior masks $V_\text{PriorMask}$ to conditionally generate future infrared sequences $V_\text{PredIR}$ and fire mask sequences $V_\text{PredMask}$. 
Output generation is controlled through modality-specific conditioning: thermal synthesis employs the prompt $\{T_\text{FireIR}\}$ for native domain prediction, while mask generation uses $\{I_\text{FireMask}, T_\text{FireMask}\}$ for domain-transformed prediction, where the image component provides essential visual reference. The coordinated training objectives are formally defined as follows:
 
(1) Task 1: Fire Infrared Modality Prediction
\begin{equation}
\begin{aligned}
V_\text{PredIR}=\text{PhysFire-WM}(&V_\text{FireIR}, V_\text{PriorMask},M,T_\text{FireIR}).
\end{aligned}
\end{equation}

(2) Task 2: Fire Mask Modality Prediction
\begin{equation}
\begin{aligned}
V_\text{PredMask}=\text{PhysFire-WM}(& V_\text{FireIR}, V_\text{PriorMask},M,  \\
& I_\text{FireMask}, T_\text{FireMask}).
\end{aligned}
\end{equation}

This dual-task framework enables parameter-efficient optimization through shared encoder components while maintaining task-specific generation capabilities via prompt-based conditioning. The CC-Train collaborative mechanism achieves balanced performance across both modalities, with comprehensive architectural comparisons provided in Supplementary Sec.~\ref{train-tasks}.
\section{Experiment}

\begin{table*}[t]
\centering
\caption{
Benchmarking fire spread prediction performance within a single region. Best and second-best results are highlighted in \textbf{bold} and \underline{underlined}, respectively. Arrows indicate the desired direction of performance ($\uparrow$ higher is better, $\downarrow$ lower is better).
}
\label{SOTA}
\renewcommand{\arraystretch}{0.9}
\resizebox{\textwidth}{!}{%
\small
\begin{tabular}{@{}c|c|cccccccc@{}}
\toprule
\multirow{2}{*}{\textbf{Category}} & \multirow{2}{*}{\textbf{Method}} & \multicolumn{4}{c}{\textbf{Fire Mask Video Quality}} & \multicolumn{4}{c}{\textbf{Fire Infrared Video Quality}} \\
\cmidrule(lr){3-6} \cmidrule(lr){7-10}
 & & \textbf{AUPRC$\uparrow$} & \textbf{F1$\uparrow$} & \textbf{IoU$\uparrow$} & \textbf{MSE$\downarrow$} & \textbf{PSNR$\uparrow$} & \textbf{SSIM$\uparrow$} & \textbf{LPIPS$\downarrow$} & \textbf{FVD$\downarrow$} \\
\midrule
\textbf{Physics-based Model} & WRF-SFIRE~\cite{mandel2014recent} & 0.75 & 0.84 & 0.73 & 0.02 & -- & -- & -- & -- \\
\midrule
\multirow{3}{*}{\textbf{Data-driven Model}} & Earthformer~\cite{gao2022earthformer} & 0.66 & 0.64 & 0.70 & 0.14 & -- & -- & -- & -- \\
 & PredRNN~\cite{wang2017predrnn} & 0.74 & 0.79 & 0.75 & 0.12 & -- & -- & -- & -- \\
 & UTAE~\cite{garnot2021panoptic}  & \underline{0.84} & 0.87 & 0.73 & \underline{0.01} & -- & -- & -- & -- \\
\midrule
\multirow{4}{*}{\textbf{Generative Model}} & MCVD~\cite{voleti2022mcvd} & 0.73 & 0.85 & 0.72 & 0.02 & 23.17 & 0.61 & 0.26 & 98.16 \\
 & STDiff~\cite{ye2024stdiff} & 0.73 & 0.86 & 0.72 & 0.01 & 24.48 & 0.67 & 0.25 & 71.54 \\
 & VDT~\cite{lu2023vdt}& 0.74 & 0.86 & 0.73 & 0.01 & 24.50 & 0.64 & 0.21 & 84.47 \\
 & DynamiCrafter~\cite{xing2024dynamicrafter} & 0.74 & 0.83 & 0.71 & 0.02 & 23.04 & 0.73 & 0.21 & 37.51 \\
\midrule
\multirow{2}{*}{\textbf{World Foundation Model}} & CogVideoX~\cite{yang2024cogvideox}& 0.79 & \underline{0.87} & \underline{0.77} & 0.02 & 22.05 & \underline{0.75} & 0.14 & 0.08 \\
 & Wan2.1-VACE-1.3B~\cite{jiang2025vace} & 0.80 & 0.86 & 0.74 & 0.02 & \underline{22.76} & 0.74 & \underline{0.12} & \underline{0.01} \\
\midrule
\textbf{Ours} & \textbf{PhysFire-WM (Our)} & \textbf{0.89} & \textbf{0.94} & \textbf{0.89} & \textbf{0.01} & \textbf{23.62} & \textbf{0.80} & \textbf{0.09} & \textbf{0.001} \\
\bottomrule
\end{tabular}%
}
\vspace{0.1cm}
\end{table*}

\begin{figure*}[t]
\centering
\includegraphics[width=0.95\linewidth]{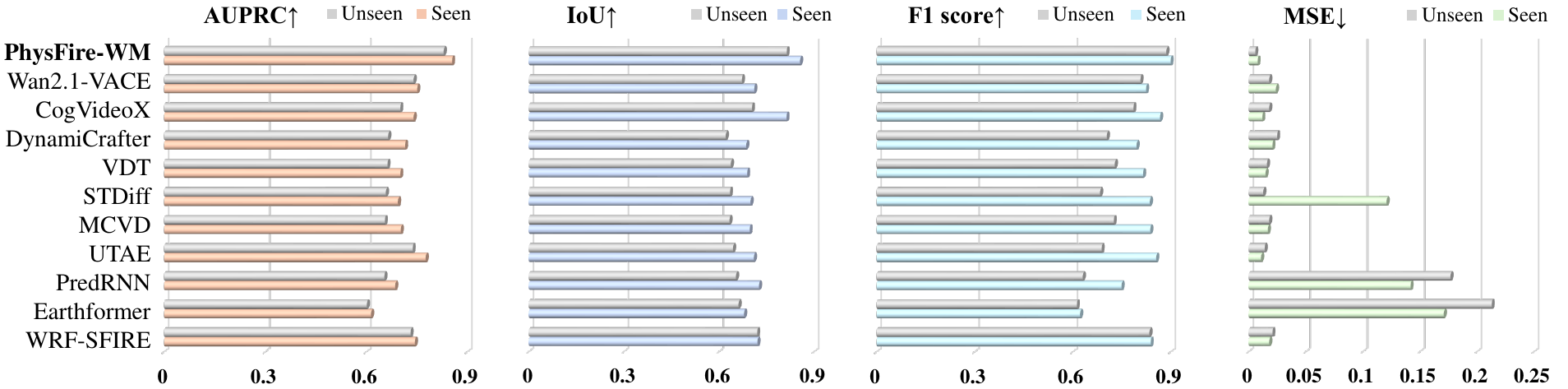}
\caption{
Model performance is evaluated across multiple regions, including both seen (training and test sets) and unseen (test set) regions.
}
\label{multi-region}
\end{figure*}

\begin{figure*}[ht]
  \centering
  \begin{subfigure}{0.49\linewidth}
    \centering
    \includegraphics[width=0.9\textwidth]{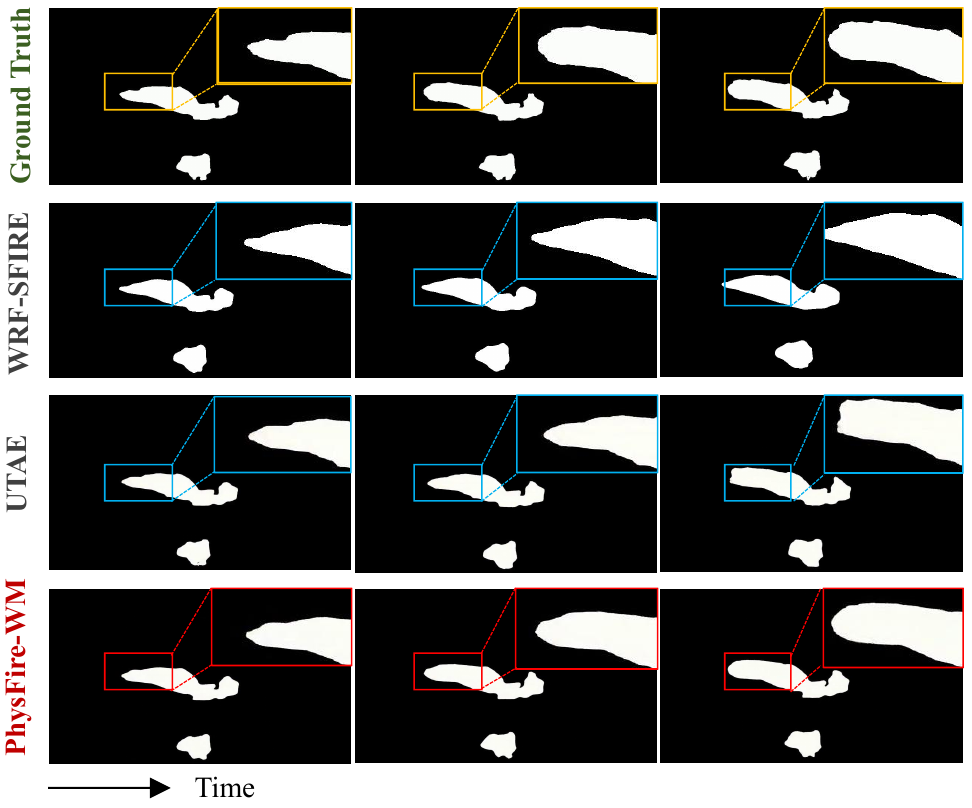}
    \caption{Mask modality results.}
    \label{visual-mask}
  \end{subfigure}
  \hfill
  \begin{subfigure}{0.49\linewidth}
    \centering
    \includegraphics[width=0.9\textwidth]{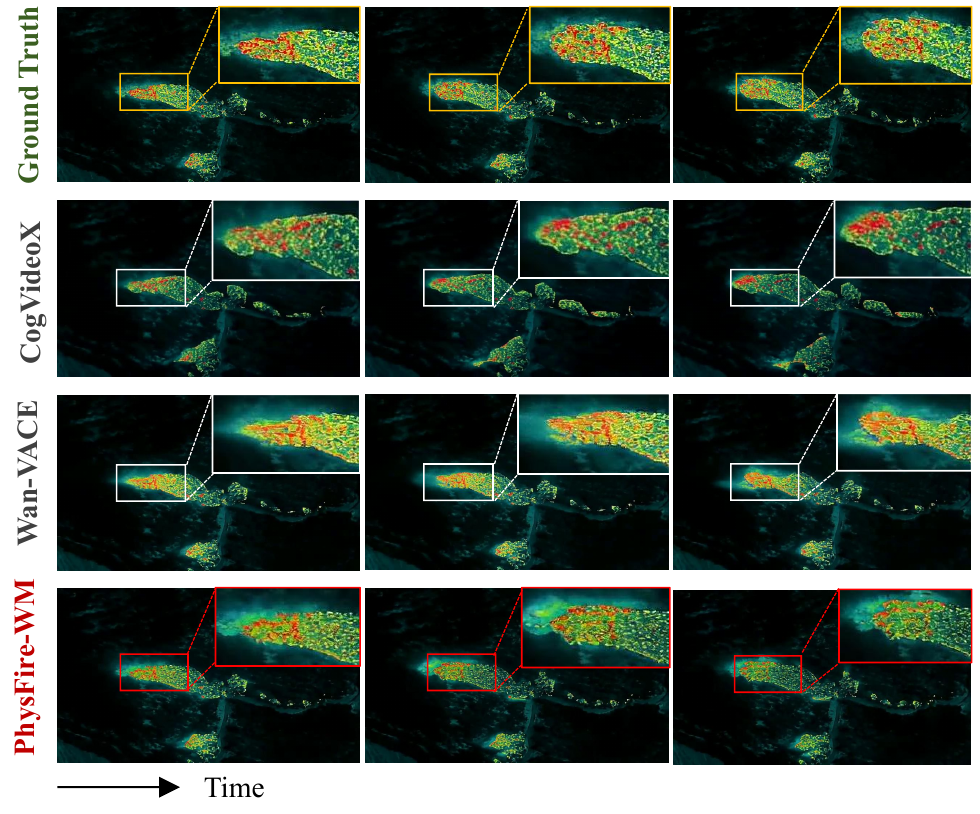}
    \caption{Infrared modality results.}
    \label{visual-IR}
  \end{subfigure}
  \caption{Visualization of Prediction Results. The enlarged view in the upper-right corner highlights the main fire spread region. (a) Mask modality prediction. (b) Infrared modality prediction.}
  \label{visual}
\end{figure*}

\subsection{Setup}
\label{setup}
\textbf{Implementation Details.} 
Our implementation builds upon the Wan2.1-VACE-1.3B architecture and its pre-trained weights~\cite{jiang2025vace}. Experimental configurations employed 3 NVIDIA RTX A6000 GPUs, with training conducted at a learning rate of 1e-4 and LoRA rank 128 to enable parameter-efficient adaptation.

\textbf{Dataset.} We constructed a fine-grained multimodal fire dataset using drones and sensors across five regions, comprising 226 spatiotemporally aligned infrared videos (480×832 resolution, 17 frames at 5-second intervals) with corresponding masks and environmental parameters. Two partitioning schemes were employed: (1) Intra-region: splitting data from individual regions; (2) Cross-region: training on majority data from three regions while testing on their remaining data plus two unseen regions. We will release the dataset.

\textbf{Baselines.} 
We evaluate ten models from four categories. Physics-based~\cite{mandel2014recent} and data-driven models~\cite{gao2022earthformer, wang2017predrnn, garnot2021panoptic} represent established wildfire prediction methods, while generative~\cite{voleti2022mcvd, ye2024stdiff, lu2023vdt, xing2024dynamicrafter} and world models~\cite{yang2024cogvideox, jiang2025vace} incorporate the latest advances, applied here to fire forecasting for the first time. For details, see Supplementary Sec.~\ref{Detail-SOTA}.

\textbf{Metrics.} 
For the wildfire spread prediction task, we employ a dual-modal evaluation framework. The accuracy of wildfire mask prediction is quantified using the Area Under the Precision-Recall Curve (AUPRC), F1-score, Intersection over Union (IoU), and Mean Squared Error (MSE). The quality of the generated infrared videos is assessed using Peak Signal-to-Noise Ratio (PSNR), Structural Similarity Index (SSIM), Learned Perceptual Image Patch Similarity (LPIPS), and Fréchet Video Distance (FVD). For details, see Supplementary Sec.~\ref{Detail-metric}.

\subsection{Quantitative Evaluation}

We evaluate each model category according to its inherent design capabilities. Physics-based and specialized data-driven wildfire models are assessed exclusively on mask prediction accuracy, while general-purpose generative and world foundation models, which are inherently designed for multimodal processing, are evaluated on both infrared video and fire mask prediction tasks. All models were uniformly conditioned on 17 historical frames to generate corresponding 17-frame future sequences.

To assess generalization capability, we employed two data partitioning schemes (Sec.~\ref{setup}). Our PhysFire-WM achieves optimal dual-modal prediction performance across both single-region (Table~\ref{SOTA}) and cross-region benchmarks (Fig.~\ref{multi-region}, with complete results in Supplementary Sec.~\ref{multi-region-metrics}).
In single-region evaluation, PhysFire-WM surpasses the second-best model across all metrics: AUPRC (+6.8\%), F1-score (+8.1\%), IoU (+15.1\%), MSE (+33.3\% reduction), PSNR (+3.7\%), SSIM (+7.1\%), LPIPS (+27.4\% improvement), and FVD (+83.3\% reduction).

\subsection{Qualitative Comparison}
For wildfire mask prediction (Fig.~\ref{visual-mask}), we compare WRF-SFIRE~\cite{mandel2014recent}, Wan2.1-VACE-1.3B~\cite{jiang2025vace}, and our PhysFire-WM. WRF-SFIRE exhibits conical propagation patterns, while Wan2.1-VACE-1.3B converges to rectangular approximations. In contrast, PhysFire-WM accurately captures the elliptical fire front morphology and maintains the closest alignment with actual spread dynamics over time.

When predicting infrared videos (Fig.~\ref{visual-IR}), baseline models exhibit clear physical inconsistencies: CogVideoX~\cite{yang2024cogvideox} introduces unrealistic artifacts, and Wan2.1-VACE-1.3B underestimates the thermal radiation. Our PhysFire-WM addresses these issues by generating radiation fields that are both visually authentic and physically consistent.

\subsection{Ablation Study}

\begin{table}[t]
\centering
\caption{Ablation study on physical prior. Best results are highlighted in \textbf{bold}. Arrows indicate the desired direction of performance ($\uparrow$ higher is better, $\downarrow$ lower is better).
}
\label{ablation-Phys}
\renewcommand{\arraystretch}{1}
\small 
\setlength{\tabcolsep}{0.1pt} 
\begin{tabular}{@{}lcccccc@{}}
\toprule
 \multirow{2}{*}{\textbf{Task}} & \multicolumn{3}{c}{Mask Quality} & \multicolumn{3}{c}{Infrared Quality} \\
\cmidrule(lr){2-4} \cmidrule(lr){5-7}
 & AUPRC$\uparrow$ & F1$\uparrow$ & IoU$\uparrow$ &  PSNR$\uparrow$ & SSIM$\uparrow$ & LPIPS$\downarrow$ \\
\midrule
\textcircled{\scriptsize 1}: Mask (w/o prior) & 0.82 & 0.86 & 0.81 &  -- & -- & --  \\
 \textcircled{\scriptsize 2}: Mask (w/ prior) & 0.85 & 0.89 & 0.83 &  -- & -- & --  \\
\textcircled{\scriptsize 3}: Infrared (w/o prior)  & -- & -- & -- & 22.76 & 0.74 & 0.12  \\
\textcircled{\scriptsize 4}: \textcircled{\scriptsize 3}+SAM2 & 0.87 & 0.91 & 0.85 & -- & -- & --  \\
\textcircled{\scriptsize 5}: Infrared (w/ prior)  & -- & -- & -- & \textbf{23.00} & \textbf{0.78} & \textbf{0.11}  \\
\textcircled{\scriptsize 6}: \textcircled{\scriptsize 5}+SAM2 & \textbf{0.88} & \textbf{0.92} & \textbf{0.86}  & -- & -- & --\\
\bottomrule
\end{tabular}
\end{table}

\begin{table}
\centering
\caption{Ablation study on CC-Train. 
"CC-" denotes results for each task when using cross-task collaborative training.
}
\label{ablation-cc}
\renewcommand{\arraystretch}{1}
\small 
\setlength{\tabcolsep}{0.1pt} 
\begin{tabular}{@{}lcccccc@{}}
\toprule
 \multirow{2}{*}{\textbf{Task}} & \multicolumn{3}{c}{Mask Quality} & \multicolumn{3}{c}{Infrared Quality} \\
\cmidrule(lr){2-4} \cmidrule(lr){5-7}
 & AUPRC$\uparrow$ & F1$\uparrow$ & IoU$\uparrow$ &  PSNR$\uparrow$ & SSIM$\uparrow$ & LPIPS$\downarrow$ \\
\midrule
 \textcircled{\scriptsize 2}: Mask (w/ prior) & 0.85 & 0.89 & 0.83 &  -- & -- & --  \\
\textcircled{\scriptsize 5}: Infrared (w/ prior)  & -- & -- & -- & 23.00 & 0.78 & 0.11  \\
\textcircled{\scriptsize 6}: \textcircled{\scriptsize 5}+SAM2 & 0.88 & 0.92 & 0.86  & -- & -- & --\\
CC-{\textcircled{\scriptsize 2}} & \textbf{0.89} & \textbf{0.94} & \textbf{0.89}  & -- & -- & --  \\
CC-{\textcircled{\scriptsize 5}} & -- & -- & --  & \textbf{23.62} & \textbf{0.80} & \textbf{0.09} \\
CC-{\textcircled{\scriptsize 6}}  & 0.88 & 0.93 & 0.88  & -- & -- & -- \\
\bottomrule
\end{tabular}
\end{table}

Ablation studies assessed the individual contributions of the physical prior and CC-Train to each modal task.

\textbf{Physical Simulator.} 
Without employing CC-Train, the configurations incorporating physical priors (\textcircled{\scriptsize 2}, \textcircled{\scriptsize 5}) outperform the prior-free settings (\textcircled{\scriptsize 1}, \textcircled{\scriptsize 3}) across all evaluation metrics in both mask and infrared prediction tasks, as shown in Table~\ref{ablation-Phys}, confirming the contribution of physical priors to dual-modal performance.
Notably, applying SAM2~\cite{ravi2024sam} post-processing to the infrared results from configuration \textcircled{\scriptsize 5} (yielding \textcircled{\scriptsize 6}) produces mask predictions that surpass all direct prediction methods (\textcircled{\scriptsize 1}, \textcircled{\scriptsize 2}).
These results validate the superiority of the "Native Domain Prediction + External Segmentation Model" approach over "Domain-Transformed Prediction".

\textbf{CC-Train.} 
CC-Train achieves dual-modal complementarity through joint optimization of mask prediction (\textcircled{\scriptsize 2}) and infrared video generation (\textcircled{\scriptsize 5}).  As summarized in Table~\ref{ablation-cc}, CC-Train-\textcircled{\scriptsize 2} outperforms \textcircled{\scriptsize 2} and CC-Train-\textcircled{\scriptsize 5} surpasses \textcircled{\scriptsize 5}, demonstrating that collaborative training effectively enhances performance compared to training each task separately.
More notably, the mask results directly output by CC-Train (CC-Train-\textcircled{\scriptsize 2}) even exceed those obtained through SAM2-dependent segmentation (CC-Train-\textcircled{\scriptsize 6}).
This demonstrates that PhysFire-WM achieves optimal performance for both fire mask and infrared predictions without relying on any external modules.
\section{Conclusion}

This paper introduces PhysFire-WM, a physics-informed world model for fire spread dynamics. By unifying physical priors with cross-task collaboration, our method achieves physical plausibility and visual fidelity in complex fire scenarios. 
Evaluated on a fine-grained multimodal fire dataset, PhysFire-WM achieves state-of-the-art results in all tasks, demonstrating robust fire spread modeling capability. Ablation studies verify that the physical prior enhances prediction plausibility while CC-Train enables synergistic performance gains beyond single-task learning, underscoring the value of physical guidance and cross-modal collaboration in disaster forecasting.

{
    \small
    \bibliographystyle{plainnat} 
    \bibliography{main}
}


\newpage

\clearpage
\setcounter{page}{1}
\maketitlesupplementary

\appendix
\section{Notation Table }
\label{sec-note}

As shown in Table~\ref{nota}, we provide a table describing the key notations used in the paper.

\begin{table*}[h]
\centering
\renewcommand{\arraystretch}{1.2}
\small
\caption{Notation table.}
\label{nota}
\begin{tabular}{c|l}
\hline
\textbf{Notation} & \textbf{Description} \\
\hline
\( g \) & Mapping Function for Fire Spread Prediciton \\
\( P_\phi \) & Physical Simulator  \\
\( G_\psi \) & Diffusion Transformer-based Generative Model \\
\( E_\eta \) & Multimodal Tokenizer \\
\( V = \{f_1, f_2, ..., f_t\} \in \mathbb{R}^{T \times H \times W \times 3} \) & Observed Video \\
\( V_{\text{FireIR}} \in \mathbb{R}^{T \times H \times W \times 3} \) & Fire Infrared Modality Video \\
\( V_{\text{PredIR}} \in \mathbb{R}^{T \times H \times W \times 3} \) & Predicted Fire Infrared Modality Video \\
\( V_{\text{FireMask}} \in \mathbb{R}^{T \times H \times W} \) & Fire Encoding Modality Video \\
\( V_{\text{PriorMask}} \in \mathbb{R}^{T \times H \times W} \) & Fire Encoding Generated by Physical Simulator  \\
\( V_{\text{PredMask}} \in \mathbb{R}^{T \times H \times W} \) & Predicted Fire Encoding Modality Video \\
\( M = \{m_1, m_2, ..., m_t\} \in \{0\}_{H \times W}, 1_{H \times W}\}^{T \times H \times W \times 3} \) & Encoding Video \\
\( X = \{x_1, x_2, ..., x_t\} \in \mathbb{R}^{(1+T/4) \times H / 8 \times W / 8} \) & The Latent Space \\
\( C_{\text{Tokens}} \) & Token Sequence Output by Multimodal Tokenizer \\
\( C_{\text{text}} \) & Text Embedding Sequence \\
\( t \) & Real-time Step of Fire Spread \\
\( n \) & Diffusion Denoising Time Step \\
\( u_n \) & Real Diffusion Velocity \\
\( \hat{u} \) & Model-predicted Diffusion Velocity \\
\( T \) & Text Prompt \\
\( I \in \mathbb{R}^{H \times W} \) & Reference Image \\
\( \mathcal{N} \in \mathbb{R}^{T \times H \times W \times 3} \) & Noisy Video \\
\( V_c \) & Reactive Frames \\
\( V_k \) & Inactive Frames \\
\( \mathcal{E} \) & Environmental Information \\
\( \mathcal{T}(p, t) \in \mathbb{R}^{T \times H \times W} \) & Fire boundary \\
\( z \in \mathbb{R}^{H \times W} \) & Terrain \\
\( F(p,t) \) & Fuel Concentration \\
\( p = (p_1, p_2) \) & 2D Coordinates \\
\( r(\mathcal{T}) \) & Burning Rate \\
\( \bar{v}(p, t) \in \mathbb{R}^{T \times H \times W} \) & Wind Velocity \\
\( c, k, \gamma, A, C, \omega \) & Coefficients \\
\( S(\mathcal{T}) \) & Heat Source Term \\
\hline
\end{tabular}
\end{table*}

\section{Design and Implementation of the PDE-Based Physical Simulator}
\label{detail-PS}
We design a Physical Simulator $P_{\phi}$ (Fig.~\ref{sim}) based on fire energy-transfer PDEs (Eq.~(\ref{PDE})), integrating its outputs as conditional priors into the world model to enforce combustion dynamics during generation. This explicit guidance significantly enhances the physical plausibility and interpretability of results. The simulator operates through three sequential components:

(1) \textit{Parametric Modeling of Combustion Source Term.} The combustion heat source term $S (\mathcal{T})$ is formulated as a convex optimization problem. To preserve physical plausibility and numerical stability, the source term is approximated as a non-negative linear combination of historical temperature fields:
\begin{equation}
S (\mathcal{T})\approx\sum_{t=1}^N \omega_t\mathcal{T}_{t}, \text{subject to } \omega_t \geq 0,\sum_t\omega_t=1.
\label{source}
\end{equation}

This constrained parametric form ensures that the estimated source term remains both physically meaningful and computationally tractable.

(2) \textit{Numerical Solution of Fire Boundary Evolution.} Using the source model above, the full energy conservation equation is expressed as:
\begin{equation}
c \frac{\partial \mathcal{T}}{\partial t}= \nabla \cdot (k \nabla \mathcal{T})(\vec{v} + \gamma \nabla z) \cdot \nabla \mathcal{T} + S (\mathcal{T})
\label{temp}
\end{equation}

Given the observed fire mask $V_\text{FireMask}$ (as $\mathcal{T}$) and environmental parameters $\mathcal{E}$ comprising terrain $z$ and wind velocity $\vec{v}$, we discretize the PDE using the finite difference method. This numerical treatment enables efficient computation of fire mask $V_\text{FireMask}$.

    
Following Eq.~(\ref{source}),~(\ref{temp}), the Physical Simulator $P_{\phi}$ integrates the observed fire mask $V_\text{FireMask}$ and environmental parameters $\mathcal{E}$ to produce a physics-driven prior mask sequence:
\begin{equation}
V_\text{PriorMask}=P_{\phi} (V_\text{FireMask},\mathcal{E}).
\end{equation}

\section{Detailed Training Task Configurations}
\label{train-tasks}

In Table~\ref{task_comparison}, we compare the training inputs and outputs of CC-Train with those of Task 1 and Task 2 (in Sec.~\ref{compoments}).

\begin{table*}[h]
\centering
\renewcommand{\arraystretch}{1.2}
\setlength{\tabcolsep}{1pt}
\caption{Comparison of training task configurations.}
\resizebox{\textwidth}{!}{
\begin{tabular}{c|c|c|c|c|c}
\hline
\multirow{2}{*}{Setting} & \multicolumn{4}{c|}{Input} & \multirow{2}{*}{Output} \\
\cline{2-5}
 & Video & Mask & Image & Text & \\
\hline
Task 1 & \( V_{\text{FireIR}} \oplus V_{\text{PriorMask}} \) & \(\{0\}\times a \oplus \{1\}\times b\) & \( I_{\text{FireIR}} \) & \begin{tabular}[c]{@{}c@{}}The fire infrared modality predicts the wildfire spread process\\ captured by an infrared camera. It incorporates prior\\ knowledge for the prediction.\end{tabular} & \begin{tabular}[c]{@{}c@{}}\( V_{\text{PredInfra}} \), \\ \( V_{\text{PredMask}} = s(V_{\text{PredIR}}) \)\end{tabular} \\
\hline
Task 2 & \( V_{\text{FireIR}} \oplus V_{\text{PriorMask}} \) & \(\{0\}\times a \oplus \{1\}\times b\) & \( I_{\text{PriorMask}} \) & \begin{tabular}[c]{@{}c@{}}The fire mask modality predicts areas of fire, with a value\\ of 1 indicating a fire and 0 indicating no fire. It incorporates\\ prior knowledge for the prediction.\end{tabular} & \( V_{\text{PredMask}} \) \\
\hline
\multirow{4}{*}{CC-Train} & \multirow{4}{*}{\( V_{\text{FireIR}} \oplus V_{\text{PriorMask}} \)} & \(\{0\}\times a \oplus \{1\}\times b\) & -- & \begin{tabular}[c]{@{}c@{}}The fire infrared modality predicts the wildfire spread process\\ captured by an infrared camera. It incorporates prior\\ knowledge for the prediction.\end{tabular} & \( V_{\text{PredIR}} \) \\
\cline{3-6}
 & & \(\{0\}\times a \oplus \{1\}\times b\) & \( I_{\text{PriorMask}} \) & \begin{tabular}[c]{@{}c@{}}The fire mask modality predicts areas of fire, with a value\\ of 1 indicating a fire and 0 indicating no fire. It incorporates\\ prior knowledge for the prediction.\end{tabular} & \( V_{\text{PredMask}} \) \\
\hline
\end{tabular}%
}
\label{task_comparison}
\end{table*}

\begin{table*}
\centering
\renewcommand{\arraystretch}{1.2}
\small
\setlength{\tabcolsep}{1pt}
\caption{Quantitative Comparison on Multi-Region Datasets. Best is highlighted in \textbf{bold}. Arrows indicate the desired direction of performance ($\uparrow$ higher is better, $\downarrow$ lower is better).}
\label{cross_region}
\begin{tabular}{c|cccc|cccc|cccc|cccc}
\hline
\multirow{3}{*}{Method} & \multicolumn{8}{c|}{Seen Region} & \multicolumn{8}{c}{Unseen Region} \\
\cline{2-17}
 & \multicolumn{4}{c|}{Mask Video Quality} & \multicolumn{4}{c|}{Infrared Video Quality} & \multicolumn{4}{c|}{Mask Video Quality} & \multicolumn{4}{c}{Infrared Video Quality} \\
\cline{2-17}
 & AUPRC$\uparrow$ & F1$\uparrow$ & IoU$\uparrow$ & MSE$\downarrow$ & PSNR$\uparrow$ & SSIM$\uparrow$ & LPIPS$\downarrow$ & FVD$\downarrow$ & AUPRC$\uparrow$ & F1$\uparrow$ & IoU$\uparrow$ & MSE$\downarrow$ & PSNR$\uparrow$ & SSIM$\uparrow$ & LPIPS$\downarrow$ & FVD$\downarrow$ \\
\hline
WRF-STRE & 0.74 & 0.84 & 0.72 & 0.02 & -- & -- & -- & -- & 0.73 & 0.84 & 0.72 & 0.02 & -- & -- & -- & -- \\
Earthformer & 0.62 & 0.62 & 0.68 & 0.17 & -- & -- & -- & -- & 0.60 & 0.61 & 0.66 & 0.21 & -- & -- & -- & -- \\
PredRNN & 0.69 & 0.75 & 0.73 & 0.14 & -- & -- & -- & -- & 0.65 & 0.63 & 0.65 & 0.18 & -- & -- & -- & -- \\
UTAE & 0.27 & 0.86 & 0.71 & 0.01 & -- & -- & -- & -- & 0.74 & 0.69 & 0.65 & 0.01 & -- & -- & -- & -- \\
MCVD & 0.70 & 0.84 & 0.70 & 0.02 & 21.78 & 0.59 & 0.39 & 117.79 & 0.66 & 0.73 & 0.63 & 0.02 & 20.86 & 0.48 & 0.41 & 147.24 \\
STDiff & 0.70 & 0.84 & 0.70 & 0.12 & 22.03 & 0.55 & 0.38 & 107.35 & 0.66 & 0.68 & 0.64 & 0.01 & 22.03 & 0.56 & 0.40 & 117.32 \\
VDT & 0.70 & 0.82 & 0.69 & 0.02 & 22.47 & 0.54 & 0.32 & 101.37 & 0.66 & 0.73 & 0.64 & 0.02 & 21.48 & 0.52 & 0.34 & 126.71 \\
DynamicCutter & 0.72 & 0.80 & 0.69 & 0.02 & 21.66 & 0.60 & 0.31 & 46.89 & 0.67 & 0.71 & 0.62 & 0.03 & 20.85 & 0.56 & 0.33 & 55.22 \\
CogVideoX & 0.74 & 0.87 & 0.81 & 0.01 & 21.13 & 0.68 & 0.19 & 1.72 & 0.70 & 0.79 & 0.70 & 0.02 & 21.61 & 0.69 & 0.12 & 0.09 \\
Wan2.1-VACE & 0.75 & 0.83 & 0.71 & 0.02 & 21.85 & 0.67 & 0.17 & 0.11 & 0.74 & 0.81 & 0.67 & 0.02 & 22.31 & 0.63 & 0.16 & 0.01 \\
\hline
\textbf{PhysFire-WM} & \textbf{0.86} & \textbf{0.92} & \textbf{0.86} & \textbf{0.01} & \textbf{23.16} & \textbf{0.74} & \textbf{0.13} & \textbf{0.02} & \textbf{0.83} & \textbf{0.89} & \textbf{0.81} & \textbf{0.01} & \textbf{23.26} & \textbf{0.71} & \textbf{0.15} & \textbf{0.00} \\
\hline
\end{tabular}
\end{table*}

\section{Details of the Compared Methods}
\label{Detail-SOTA}

We benchmark our approach against 10 representative baseline methods: 

\begin{itemize}
\item \textbf{WRF-SFIRE}~\cite{mandel2014recent}: A physics-based simulation method coupling the Weather Research and Forecasting model with the fire spread model SFIRE. It initiates simulation from observed fire boundaries and provides the governing equations for our physical simulator.
\item \textbf{Earthformer}~\cite{gao2022earthformer}: Transformer-based architecture with specialized spatiotemporal attention for earth system forecasting tasks including wildfire and precipitation prediction.

\item \textbf{PredRNN}~\cite{wang2017predrnn}: RNN-based model featuring spatiotemporal LSTM units to simultaneously capture spatial and temporal dynamics for video prediction.

\item \textbf{UTAE}~\cite{garnot2021panoptic}: U-Net-based encoder-decoder framework incorporating temporal self-attention for multi-scale spatiotemporal feature extraction in segmentation and forecasting.

\item \textbf{MCVD}~\cite{voleti2022mcvd}: Conditional denoising diffusion model supporting mask-controlled prediction and interpolation tasks, employing U-Net as the denoising network.

\item \textbf{STDiff}~\cite{ye2024stdiff}: Video diffusion model that disentangles motion and content for autoregressive generation, using U-Net as the denoising network.

\item \textbf{VDT}~\cite{lu2023vdt}: Diffusion transformer framework with unified spatiotemporal mask modeling for video prediction, utilizing Transformer as the denoising network.

\item \textbf{DynamicCrafter}~\cite{xing2024dynamicrafter}: Text-to-video diffusion approach leveraging motion priors for image-conditioned generation, with 3D U-Net as the denoising network.

\item \textbf{CogVideoX}~\cite{yang2024cogvideox}: Diffusion transformer-based text-to-video model generating coherent long videos, lacking video conditioning channels, using Transformer for denoising.

\item \textbf{Wan2.1-VACE-1.3B}~\cite{jiang2025vace}: General-purpose video diffusion transformer with context adapter for multimodal conditioning, serving as our base architecture with Transformer denoising.

\end{itemize}

\section{Description of Evaluation Metrics}
\label{Detail-metric}

To evaluate both video generation quality and fire mask segmentation accuracy, we adopt a total of eight metrics. Specifically, we use four standard metrics for assessing video generation: PSNR, SSIM, LPIPS, and FVD; and four metrics for evaluating fire mask segmentation: AUPRC, F1 Score, IoU, and MSE. Definitions are provided below for each metric.

\textit{PSNR (Peak Signal-to-Noise Ratio)} quantifies the fidelity between predicted and ground-truth video frames. Higher values indicate better reconstruction quality:
\begin{equation}
\mathrm{PSNR} = 10 \cdot \log_{10} \left( \frac{\mathrm{MAX}^2}{\mathrm{MSE}} \right)
\end{equation}
where $\mathrm{MAX}$ is the maximum possible pixel value (e.g., 255), and $\mathrm{MSE}$ denotes mean squared error between corresponding frames.

\textit{SSIM (Structural Similarity Index)} evaluates perceptual similarity between frames, taking into account luminance, contrast, and structural information. Higher scores imply better perceptual quality:
\begin{equation}
\mathrm{SSIM}(x, y) = \frac{(2\mu_x\mu_y + C_1)(2\sigma_{xy} + C_2)}{(\mu_x^2 + \mu_y^2 + C_1)(\sigma_x^2 + \sigma_y^2 + C_2)}
\end{equation}
where $\mu_x$, $\mu_y$, $\sigma_x^2$, $\sigma_y^2$, and $\sigma_{xy}$ represent the means, variances, and covariance of images $x$ and $y$, respectively. $C_1$ and $C_2$ are small constants to avoid division by zero.


\textit{LPIPS (Learned Perceptual Image Patch Similarity)} measures perceptual similarity between images by comparing deep feature activations extracted from a pretrained neural network. Lower LPIPS values indicate higher perceptual similarity. Formally, given two images \(x\) and \(\hat{x}\), LPIPS is computed as
\begin{equation}
\mathrm{LPIPS}(x, \hat{x}) = \sum_{l} \frac{1}{H_l W_l} \sum_{h=1}^{H_l} \sum_{w=1}^{W_l} w_l \cdot \left\| \hat{y}^l_{h,w} - y^l_{h,w} \right\|_2^2,
\end{equation}
where \(y^l\) and \(\hat{y}^l\) are the feature maps at layer \(l\) of the network for images \(x\) and \(\hat{x}\), respectively; \(H_l, W_l\) denote the spatial dimensions of the feature map; and \(w_l\) are learned scalar weights for each layer. Since LPIPS relies on a pretrained network, it does not have a closed-form expression.

\textit{FVD (Fréchet Video Distance)} measures the distributional distance between generated and real video feature distributions. Lower values reflect better realism:
\begin{equation}
\mathrm{FVD} = \|\mu_r - \mu_g\|^2 + \mathrm{Tr}\left( \Sigma_r + \Sigma_g - 2(\Sigma_r \Sigma_g)^{1/2} \right)
\end{equation}
where $(\mu_r, \Sigma_r)$ and $(\mu_g, \Sigma_g)$ are the means and covariances of feature representations from real and generated videos.

\textit{AUPRC (Area Under the Precision-Recall Curve)} evaluates segmentation performance by summarizing the trade-off between precision and recall over varying classification thresholds. A higher AUPRC value indicates better overall segmentation quality. 
Formally, given precision \( P(r) \) as a function of recall \( r \), the AUPRC is defined as the integral:
\begin{equation}
\mathrm{AUPRC} = \int_0^1 P(r) \, dr,
\end{equation}
where precision and recall are computed as
\begin{equation}
\text{Precision} = \frac{TP}{TP + FP}, \quad \text{Recall} = \frac{TP}{TP + FN},
\end{equation}
with \(TP\), \(FP\), and \(FN\) denoting true positives, false positives, and false negatives, respectively.

\textit{F1 Score} captures the harmonic mean of precision and recall, providing a balanced measure of accuracy:
\begin{equation}
\mathrm{F1\ Score} = \frac{2 \cdot \mathrm{Precision} \cdot \mathrm{Recall}}{\mathrm{Precision} + \mathrm{Recall}}
\end{equation}
Higher values represent more accurate fire mask predictions.

\textit{IoU (Intersection over Union)} measures the overlap between predicted and ground-truth segmentation masks:
\begin{equation}
\mathrm{IoU} = \frac{|P \cap G|}{|P \cup G|}
\end{equation}
where $P$ and $G$ are the predicted and ground-truth mask regions, respectively. Larger values indicate better alignment.

\textit{MSE (Mean Squared Error)} quantifies the average squared difference between predicted and ground-truth mask pixels:
\begin{equation}
\mathrm{MSE} = \frac{1}{N} \sum_{i=1}^{N} (x_i - y_i)^2
\end{equation}
where $x_i$ and $y_i$ denote pixel values of the predicted and ground-truth masks. Lower values indicate more accurate segmentation.

\section{Model Performance Across Multi-Region Datasets}
\label{multi-region-metrics}

As shown in Table~\ref{cross_region}, we evaluated our method's performance on a dataset comprising multiple regions. Our approach achieved optimal results across all metrics, demonstrating strong generalization capability.

\end{document}